\documentclass[conference]{IEEEtran}
\IEEEoverridecommandlockouts

\usepackage{cite}
\usepackage{amsmath,amssymb,amsfonts}
\usepackage{algorithmic}
\usepackage{graphicx}
\usepackage{textcomp}
\usepackage{xcolor}
\usepackage{graphicx}
\usepackage{bm}
\usepackage{multirow,multicol}
\usepackage{makecell}
\usepackage{bbding}
\usepackage{threeparttable}

\def\BibTeX{{\rm B\kern-.05em{\sc i\kern-.025em b}\kern-.08em
    T\kern-.1667em\lower.7ex\hbox{E}\kern-.125emX}}
\begin{document}

\title{Transducer-Llama: Integrating LLMs into Streamable Transducer-based Speech Recognition\\
\thanks{\IEEEauthorrefmark{3}Corresponding author.}
\thanks{Work done while Keqi Deng was an intern at Meta AI.}
}

\author{\IEEEauthorblockN{Keqi Deng\IEEEauthorrefmark{1}\IEEEauthorrefmark{2}, Jinxi Guo\IEEEauthorrefmark{1}\IEEEauthorrefmark{3}, Yingyi Ma\IEEEauthorrefmark{1}, Niko Moritz\IEEEauthorrefmark{1}, Philip C. Woodland\IEEEauthorrefmark{2}, Ozlem Kalinli\IEEEauthorrefmark{1}, Mike Seltzer\IEEEauthorrefmark{1}}
\IEEEauthorblockA{\IEEEauthorrefmark{1}Meta AI, USA\\
\{keqi, jinxiguo, yingyima, nmoritz, okalinli, mikeseltzer\}@meta.com}
\IEEEauthorblockA{\IEEEauthorrefmark{2}Department of Engineering, University of Cambridge, UK\\
\{kd502, pw117\}@cam.ac.uk}}

\maketitle

\begin{abstract}
While large language models (LLMs) have been applied to automatic speech recognition (ASR), the task of making the model streamable remains a challenge.
This paper proposes a novel model architecture, Transducer-Llama, that integrates LLMs into a Factorized Transducer (FT) model, naturally enabling streaming capabilities. 
Furthermore, given that the large vocabulary of LLMs can cause data sparsity issue and increased training costs for spoken language systems, this paper introduces an efficient vocabulary adaptation technique to align LLMs with speech system vocabularies.
The results show that directly optimizing the FT model with a strong pre-trained LLM-based predictor using the RNN-T loss yields some but limited improvements over a smaller pre-trained LM predictor. Therefore, this paper proposes a weak-to-strong LM swap strategy, using a weak LM predictor during RNN-T loss training and then replacing it with a strong LLM.
After LM replacement, the minimum word error rate (MWER) loss is employed to finetune the integration of the LLM predictor with the Transducer-Llama model.
Experiments on the LibriSpeech and large-scale multi-lingual LibriSpeech corpora show that the proposed streaming Transducer-Llama approach gave a 17\% relative WER reduction (WERR) over a strong FT baseline and a 32\% WERR over an RNN-T baseline.

\end{abstract}


\section{Introduction}
\label{sec:intro}
End-to-end (E2E) automatic speech recognition (ASR) simplifies conventional pipeline methods and directly transcribes speech into text \cite{graves2006connectionist, Graves2012SequenceTW}. In many real-world scenarios, streaming ASR is needed for low latency.
Many E2E approaches \cite{graves2006connectionist, Graves2012SequenceTW, deng2024label} have been developed for such online applications, among which the recurrent neural network transducer (RNN-T) \cite{Graves2012SequenceTW} is widely used for streaming operation. However, while the prediction network of the RNN-T has a similar structure to a language model (LM), it does not perform as an explicit LM \cite{deng2024label, 9054419, Chen2021FactorizedNT}, which makes it hard to incorporate pre-trained LMs \cite{deng2024decoupled}.
Several papers \cite{Chen2021FactorizedNT, meng2023modular, deng2024decoupled, guo2024effective} have tried to separate the LM component from the neural transducer, e.g. the non-blank predictor in the Factorized Transducer (FT) \cite{Chen2021FactorizedNT, guo2024effective}. However, a major challenge is that, even with a modularized internal LM component, use of a strong internal LM provides limited ASR improvement on general datasets \cite{meng2023modular, zhao2023fast}.  Recently, \cite{guo2024effective} proposed an effective internal LM fusion and training strategy that greatly improves the FT model performance. 
 
Text-based large LMs (LLMs) have achieved great success \cite{brown2020language, touvron2023llama2, dubey2024llama, ouyang2022training}. 
Several recent studies have focused on extending text-based LLMs to handle speech input (speech LLMs), in which speech prompts are prepended to the text sequence and the LLMs are conditioned on the input speech \cite{10447605, deng2024wav2prompt}. However, this decoder-only architecture cannot naturally handle streaming as the speech prompts are prepended before text input \cite{chen2024bestow}. In addition, text-based LLMs operate on discrete text units, in which a tokenizer maps the raw text into a token sequence, and LLMs are limited to the vocabulary they were trained on \cite{minixhofer2024zero}. However, this restricts the flexibility when applying LLMs to speech tasks. For example, the vocabulary size of LLMs can be too large to be used for ASR system training \cite{9746316}. In addition, the LLM tokenizer may not be optimal for domains for which it wasn't designed \cite{minixhofer2024zero, dagan2024getting}.

In order to efficiently integrate LLMs into streaming ASR systems, this paper proposes a novel architecture, Transducer-Llama, which is based on the recently proposed FT model \cite{guo2024effective}, using an LLM as a non-blank predictor. To avoid the data sparsity and increased training costs caused by the large LLM vocabulary, this paper introduces an efficient vocabulary adaptation technique, aligning the LLM with a specialized ASR vocabulary. 
Preliminary experiments show that, compared to a weaker pre-trained LM predictor, using a strong LLM-based non-blank predictor leads to only minor ASR improvements after RNN-T loss training.
To address this, we propose a weak-to-strong LM swap strategy, using a weak LM (e.g. stateless predictor) during RNN-T loss training and then replacing it with a strong LLM. After LM replacement, the internal LM-aware minimum word error rate (MWER) loss \cite{guo2020mwer, guo2024effective} is applied to further finetune the Transducer-Llama model in order to optimize the integration of the LLM predictor.
Therefore, Transducer-Llama provides a framework that naturally applies LLMs to online ASR and enables highly efficient training.
Experiments on LibriSpeech \cite{7178964} and Multi-lingual LibriSpeech (MLS) \cite{pratap20_interspeech} show the Transducer-Llama can effectively incorporate LLMs to boost ASR accuracy, and also demonstrate superior performance over speech LLMs.


\begin{figure}[t]
    \centering
    \includegraphics[width=83mm]{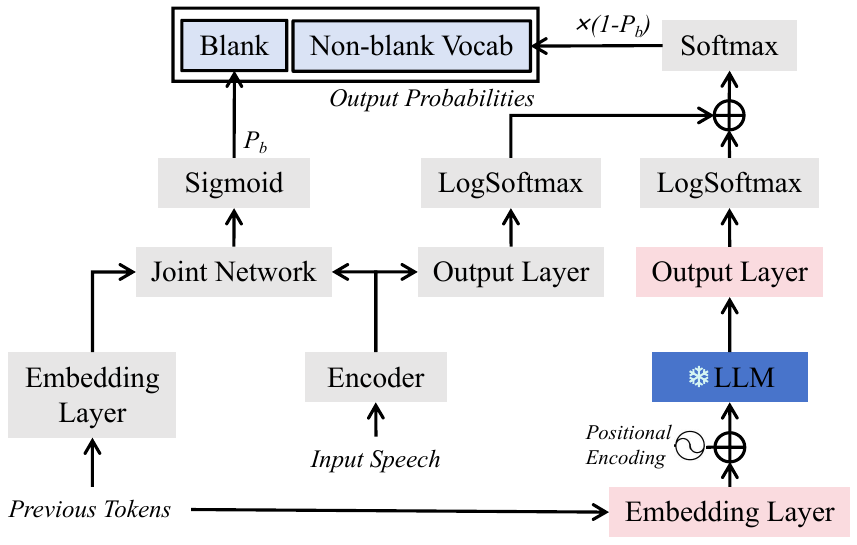}
    \vspace{-0.3cm}
    \caption{Illustration of the Transducer-Llama framework. $\bm{\oplus}$ denotes addition. LLM parameters are fixed (frozen). The output and embedding layers in red are initialised based on the proposed vocabulary adaptation method.}
    \vspace{-0.1cm}
    \label{llm-t}
\end{figure}

\section{Related Work}
\label{related}
Given the success of LLMs and the importance of low-latency responses in ASR, recent work has begun to explore online LLM-based speech systems. However, the speech input is always prepended before text input as a prompt, making online operation challenging \cite{chen2024bestow}.
Hence, online modifications must be made based on known speech-text alignments in order to restrict the speech visible for each text token. \cite{seide2024speech} uses an external CTC model to provide the alignment, which breaks the flat-start advantage of E2E models. \cite{tsunoo2024decoder} obtains the speech-text alignment on the fly at training via a CTC greedy search, but this reduces training efficiency. \cite{chen2024bestow} instead uses a Wait-k \cite{Ma2020SimulMTTS} strategy to provide a hard-coded alignment, however, as noted in \cite{deng2024lst, chen2024bestow}, this strategy is unsuitable for low-latency scenarios. In this paper, the proposed Transducer-Llama provides a natural solution to support LLM-based online ASR, which can be efficiently trained from a flat-start.

Several papers have separated the internal LM from the neural transducer \cite{Chen2021FactorizedNT, meng2023modular, deng2024decoupled, guo2024effective}. This paper uses the modified FT structure based on \cite{guo2024effective}, which can be referred to Fig.~\ref{llm-t} (gray parts), except for the LLM part.
Separate predictors are used, i.e. a blank predictor and non-blank predictor (internal LM), the non-blank probability $P_{\text{nb}}$ is obtained from the sum of the acoustic encoder (${\rm log}P_{\text{ac}}$) and internal LM log-probabilities (${\rm log}P_{\text{ilm}}$). 
The outputs of the encoder and blank predictor (e.g. gray embedding layer in Fig.~\ref{llm-t})
are combined in joint network to produce a single-dimension logit, which is then passed through a sigmoid function to compute blank probability $P_{\text{b}}$.
The non-blank probability $P_{\text{nb}}$ is normalized to decouple it from $P_{\text{b}}$: $P_{\text{nb}}=(1-P_{\text{b}})({\rm softmax}({\rm log}P_{\text{ac}}+{\rm log}P_{\text{ilm}})$. 
During decoding, \cite{guo2024effective} uses a modified score for non-blank tokens: 
\begin{equation}
    {\rm log}((1-P_{\text{b}})({\rm softmax}({\rm log}P_{\text{ac}}+\alpha{\rm log}P_{\text{ilm}}))+\beta{\rm log}P_{\text{ilm}} \label{decode}
\end{equation}
where $\alpha$ and $\beta$ are hyper-perameters.

\section{Transducer-Llama}
The proposed Transducer-Llama, as shown in Fig.~\ref{llm-t}, uses the FT architecture and incorporates LLMs as the non-blank predictor at decoding to model causal dependencies, which is the main difference from FT.
An embedding layer is used as the blank predictor.
The vocabulary adaptation technique, weak-to-strong LM swap strategy followed by MWER training are employed to make Transducer-Llama training efficient and fully utilise the LLM to boost ASR performance.

\subsection{LLM Vocabulary Adaptation}
LLMs have large vocabularies not designed for speech systems. This makes training expensive, especially for the RNN-T loss. Therefore, this paper designs a vocabulary adaptation method to adapt the LLM to the ASR system vocabulary. The ASR tokenizer is trained on transcripts from ASR training data, which is typically much smaller than the LLM vocabulary, helping alleviate data sparsity. To achieve efficient training, the LLM can be fixed at training as shown in Fig.~\ref{llm-t} while only updating the embedding and output layers. 

To align LLMs to the ASR vocabulary, new embedding and output layers need to be employed, since their weight matrices are tied to the vocabulary size, while the Transformer layers of the LLM are kept fixed. To efficiently adapt the vocabulary,
while leveraging the original LLM information, this paper initializes these weight matrices (highlighted in red in Fig.~\ref{llm-t}) based on the original LLM vocabulary, inspired by \cite{DBLP:conf/emnlp/DoblerM23, DBLP:conf/emnlp/GeeZRT22, DBLP:conf/naacl/MinixhoferPR22, chen2024model}.

Suppose the weight matrix in the embedding or output layer is $\textbf{W}^{\text{new}} \in \mathbb{R}^{|V|\cdot d}$,
where $V$ is the ASR vocabulary with size $|V|$ and $d$ is the embedding dimension. Denote the LLM vocabulary as $U$ and the corresponding weight matrix as $\textbf{W}^{\text{llm}} \in \mathbb{R}^{|U|\cdot d}$. 
For each token $t_i\in V$, if $t_i\in V\cap U$, then $\bm{W}^{\text{new}}_{i} = \bm{W}^{\text{llm}}_{j}$, where $j$ is the corresponding index of $t_i$ in the original LLM vocabulary. 
If $t_i\notin U$, then $t_i$ is tokenized using the LLM tokenizer, and $\bm{W}^{\text{new}}_{i}$ is initialized as the average of the corresponding embeddings. If it cannot be decomposed, it is initialized randomly. Note the newly-trained weight matrices of the embedding and output layer are distinct in this paper.

\subsection{Weak-to-Strong LM Swap}
Preliminary experiments showed that when training with the RNN-T loss, introducing a stronger LM as the non-blank predictor results in relatively limited improvements in ASR performance, consistent with previous work \cite{meng2023modular, zhao2023fast}. To address this challenge, we propose a weak-to-strong LM swap strategy, using a weak LM during training and replaced by a strong LLM at decoding. This prevents the model from overly relying on the strong LM accurate predictions during training and ensures the acoustic information is fully optimized and utilized. Additionally, this strategy greatly speeds up the training process. 
In this paper, an embedding layer is used as the weak LM, functioning similarly to a bigram LM and also called a stateless non-blank predictor.

\subsection{Proposed Training and Decoding}
We first train a FT model with the weak LM non-blank predictor from scratch. The RNN-T loss is used together with an additional internal LM (ILM) loss \cite{guo2024effective} to train the internal LM. Once the model has converged, the small internal LM is replaced by an LLM, thereby transforming the FT model into Transducer-Llama. Transducer-Llama can be used for decoding following Eq.~\ref{decode}, in which the LLM predicted probabilities $P_{\text{LLM}}$ is considered as $P_{\text{ilm}}$.
This weak-to-strong LM swap strategy forces the encoder to fully utilize acoustic information to improve the accuracy of $P_{\text{nb}}$. Compared to integrating LLM directly at RNN-T training stage, it can fully leverage the capabilities of the LLM to improve ASR performance while also offering fast training speed.


After the LM swap, the internal LM-aware MWER loss (proposed in \cite{guo2024effective}) is used to train the Transducer-Llama, which directly optimizes the LLM predictor integration with the word-level edit distance and performs sequence discriminative training. The N-best hypotheses are generated according to Eq.~\ref{decode}, where the weights $\alpha$ and $\beta$ applied during MWER training are also used during inference. 
It's shown in Section IV that MWER loss can more effectively leverages strong LMs to improve ASR performance, in comparison to RNN-T loss.

With its modular structure, weak-to-strong LM swap following by MWER training, Transducer-Llama shares many similarities with the conventional neural network-hidden Markov model (HMM) hybrid sequence discriminative training 
strategy \cite{vesely2013sequence, 5743665, schluter1999interdependence} or external LM fusion \cite{8682490}.
\section{Experiments}
\label{sec:typestyle}
\subsection{Corpus}
Experiments were conducted using the LibriSpeech \cite{7178964} and multi-lingual LibriSpeech (MLS) \cite{pratap20_interspeech} corpora. Four languages, English (en), French (fr), Italian (it) and Dutch
(nl), from the MLS dataset were used as the multi-lingual training data, with respective
training audio size of 44.7k hrs, 1.1k hrs, 0.2k hrs, 1.6k hrs.
The LibriSpeech LM corpus (800M words) and training data transcripts were used to train the LMs for LibriSpeech experiments. For MLS, the LibriSpeech LM data (En), and the French (146M words), Italian (41M words), and Dutch (46M words) text data from the MLS LM corpus, were used. This multilingual text data, along with training data transcripts, was employed to train the LMs for MLS.

\subsection{Model Descriptions}
\label{setup}
All ASR systems used a vocabulary of 5000 tokens along with the blank token for both LibriSpeech and MLS data. For LibriSpeech, the streaming encoder used a 20-layer Emformer \cite{shi2021emformer} with a 160~ms segment size, 512 attention dimensions, 2048 feed-forward dimensions, and 8 heads (63M parameters). With the same attention configurations, a 30-layer streaming Conformer (190M) \cite{gulati20_interspeech} was used for MLS data, in which a chunk-based mask was implemented with a 320~ms average latency. Model input used 80~d filterbank features with a 10~ms frame rate were used with $1/4$ down-sampling.

The standard RNN-T model and FT model from \cite{guo2024effective} were built to compare with Transducer-Llama. All of these used the same encoder. The standard RNN-T had a predictor consisting of two LSTM layers with 2048 hidden dimensions. With the same structure, the non-blank predictor of the FT baseline was pre-trained on text-only data and kept fixed during ASR training. The values of $\alpha$ and $\beta$ in Eq.~\ref{decode} were set to 0.6 following \cite{guo2024effective}.
Aside from the non-blank predictor, Transducer-Llama shared the same settings as the FT baseline. During training, an embedding layer was used as the non-blank predictor.
Aside from the LSTM LM used by the FT baseline, Llama2-0.5b \cite{touvron2023llama2} and Llama3-8b \cite{dubey2024llama} were fine-tuned on the same text data using the proposed vocabulary adaptation, in which the LLM Transformer layers are fixed. 
This paper also evaluated the case when the Llama2-0.5b Transformer layers were also updated,
denoted as fully fine-tuned Llama2 (FFT Llama2).
The LLM replaced the used LM of the Transducer-Llama as the 
non-blank predictor after RNN loss training. 

For LibriSpeech, the ASR models were trained for 40 epochs, while for MLS, the ASR models were trained for 200k steps with a 5.3m total batch size. MWER training ran for up to a few thousand steps. The LM was trained for up to 40 epochs, while the vocabulary adaptation training for LLMs often stopped early after a few epochs.
At decoding, the beam search size was 10.

\subsection{Experimental Results}
The proposed Transducer-Llama was evaluated on both LibriSpeech and MLS data. Ablation studies were conducted to verify the effectiveness of the vocabulary adaptation technique and the weak-to-strong LM swap.

\begin{table}[t]
    \caption{WER on LibriSpeech test sets for streaming models (160ms segment size). FFT Llama2 is fully fine-tuned (FFT) Llama2.}
  \label{tab:ls960}
  \centering
  \setlength{\tabcolsep}{3.0mm}
  \renewcommand\arraystretch{1.25}
  \begin{tabular}{l c c}
    \Xhline{3\arrayrulewidth}
    Online Models&Test-clean &Test-other\\
    \hline
    RNN-T&3.65&8.96\\
    Factorized Transducer \cite{guo2024effective} &2.97&7.77\\
    Transducer-Llama\\
    {\ }w/ LSTM LM non-blank predictor&2.86&7.33\\
    {\ }w/ Llama2 non-blank predictor&2.76&7.07\\
    {\ }w/ FFT Llama2 non-blank predictor& 2.54&6.59 \\
    {\ }w/ Llama3 non-blank predictor& 2.47&6.53\\
    \Xhline{3\arrayrulewidth}
  \end{tabular}
\end{table}
\begin{table}[t]
    \caption{Ablation studies on the weak-to-strong LM swap method using LibriSpeech data.}
  \label{ablation:s2l}
  \centering
  \setlength{\tabcolsep}{0.9mm}
  \renewcommand\arraystretch{1.25}
  \begin{tabular}{l |l|c|c| c}
    \Xhline{3\arrayrulewidth}
    Train-time & Test-time&\multirow{2}{*}{MWER}&\multirow{2}{*}{Test-clean} &\multirow{2}{*}{Test-other}\\
    Non-blank Predictor&Non-blank Predictor&&&\\
    \hline
    LSTM LM&LSTM LM&\XSolidBrush&3.11&7.83\\
    LSTM LM&LSTM LM&\Checkmark&2.97&7.77\\
    Stateless&LSTM LM&\XSolidBrush&3.20&7.59\\
    Stateless&LSTM LM&\Checkmark&\textbf{2.86}&\textbf{7.33}\\
    \hline
    LSTM LM&Llama2&\XSolidBrush&2.95&7.45\\
    LSTM LM&Llama2&\Checkmark&2.82&7.41\\
    Stateless&Llama2&\XSolidBrush&3.39&7.47\\
    Stateless&Llama2&\Checkmark&\textbf{2.76}&\textbf{7.07}\\
    Llama2&Llama2&\XSolidBrush&3.03&7.63\\
    Llama2&Llama2&\Checkmark&2.90&7.41\\
    \hline
    LSTM LM&FFT Llama2&\XSolidBrush&2.67&6.95\\
    LSTM LM&FFT Llama2&\Checkmark&2.61&6.94\\
    Stateless&FFT Llama2&\XSolidBrush&3.00&6.91\\
    Stateless&FFT Llama2&\Checkmark&\textbf{2.54}&\textbf{6.59}\\
    FFT Llama2&FFT Llama2&\XSolidBrush&2.86&7.51\\
    FFT Llama2&FFT Llama2&\Checkmark&2.72&7.25\\
    \Xhline{3\arrayrulewidth}
  \end{tabular}
\end{table}

\subsubsection{LibriSpeech Main Results}

As shown in Table~\ref{tab:ls960}, the FT baseline outperformed the RNN-T baseline, which is consistent with \cite{guo2024effective}.
In addition, the proposed Transducer-Llama approach provides a framework that fully leverages LMs to enhance ASR performance. As more powerful LMs are used, the WER of the LM-Transducer continues to decrease. Compared to the strong FT baseline, up to 16.8\% and 16.0\% relative WER reduction (WERR) were achieved on test-clean and test-other, respectively.
This also shows that with our vocabulary adaptation, the LLM can efficiently adapt to the ASR vocabulary while maintaining strong performance, offering advantages for practical ASR deployments. In addition,
Llama3 slightly outperforms the fully fine-tuned (FFT) Llama2,
but by keeping the Transformer layers fixed, it retains the potential for a broader range of downstream tasks.
Moreover, the ASR improvement (6\% WERR) of the Transducer-Llama with the LSTM LM as a non-blank predictor, compared to the FT baseline, highlights the advantages of the weak-to-strong LM swap strategy.
Detailed ablation studies are given in Sec.~\ref{small-to-large}.

\begin{table}[t]
    \caption{Ablation studies on vocabulary adaptation using LibriSpeech with Llama3 as the non-blank predictor. Weak-to-Strong LM swap and MWER training were not used.
    Llama3 tokenize has 128k vocabulary size compared to 5k for ASR.}
  \label{ablation:ls960}
  \centering
  \setlength{\tabcolsep}{3.0mm}
  \renewcommand\arraystretch{1.25}
  \begin{tabular}{l |c| c|c c}
    \Xhline{3\arrayrulewidth}
    \multirow{2}{*}{Online Models}&\multirow{2}{*}{Tokenizer}&Train&\multicolumn{2}{c}{Test}\\
    &&Speed&clean&other\\
    \hline
    Transducer-Llama&{Llama3}&{1}&{3.02}&{7.44}\\
    \hline
    Transducer-Llama&{ASR}&{$\times$8.05}&{2.76}&{7.36}\\
    \Xhline{3\arrayrulewidth}
  \end{tabular}
\end{table}

\begin{table}[t]
    \caption{WER on MLS test sets for different models. The last column shows the average WER on the 4 languages. LSTM predictor has 20M parameters, Llama2 has 0.5B parameters, Llama3 has 8B parameters. Our built models are multi-lingual ASR.}
  \label{tab:mls}
  \centering
  \setlength{\tabcolsep}{1.15mm}
  \renewcommand\arraystretch{1.25}
  \begin{tabular}{l c cccc}
    \Xhline{3\arrayrulewidth}
    \textbf{Online} Models&en &fr&it&nl& Avg\\
    \hline
    Factorized Transducer \cite{guo2024effective} &8.19&7.33&14.02&13.37&10.73\\
    Transducer-Llama\\
    {\ }w/ LSTM LM non-blank predictor&8.26&6.28&12.30&12.39&9.80\\
    {\ }w/ FFT Llama2 non-blank predictor&7.57&5.80&11.59&12.17&9.28\\
    {\ }w/ Llama3 non-blank predictor&7.35&5.79&11.76&12.05&9.24\\
    \hline
    \textbf{Offline} Mono-lingual CTC w/ LM\cite{pratap20_interspeech}&5.9&5.6&10.5&12.0&8.50\\
    \textbf{Offline} Speech LLM\cite{10447605}&6.2&5.5&10.8&11.3&8.45\\
    \textbf{Offline} Transducer-Llama\\
    {\ }w/ FFT Llama2 non-blank predictor&5.76&5.00&10.38&10.28&7.86\\
    {\ }w/ Llama3 non-blank predictor&5.59&4.96&10.43&10.22&7.80\\
    \Xhline{3\arrayrulewidth}
  \end{tabular}
\end{table}

\subsubsection{Ablation Studies on Weak-to-Strong LM Swap}
\label{small-to-large}

As shown in Table~\ref{ablation:s2l}, even with the ILM fusion strategy (Eq.~\ref{decode}), integrating stronger LLMs 
during RNN-T loss training yields only minor ASR improvements.
For example,
compared to using an LSTM LM as the non-blank predictor, employing stronger fully fine-tuned (FFT) Llama2 at training results in only 4\% WERR on test-other.
However, when the weak-to-strong LM swap strategy is used, e.g. using an LSTM LM during training and a FFT Llama2 during decoding, the ASR WERR greatly increased and gave a 14\% WERR on test-clean and 11\% WERR and test-other.
While using a stateless non-blank predictor during RNN-T loss training performs slightly worse than when using an LSTM LM after swapping to an LLM on test-clean, it achieves the best results after MWER training. The use of a stateless predictor, being even weaker than an LSTM, has a larger gap to the LLM, so after swap, it does not surpass the performance of LSTM-trained models. MWER training is designed to address this issue and optimizes the LLM predictor integration, which is especially effective for the stateless non-blank predictor training case (up to 18.6\% WERR).
Using a weak LM as the non-blank predictor during RNN-T loss training speeds up the training process and prevents the model from relying on the accurate predictions of a strong LM, ensuring the acoustic encoder is properly trained.

\subsubsection{Ablation Studies on Vocabulary Adaptation}

Table~\ref{ablation:ls960} compares the Transducer-Llama performance using the ASR vocabulary versus its original Llama3 tokenizer. 
The output of the neural transducer is known to be memory-intensive because it is a 4-dimensional tensor, including the vocabulary dimension. Given the vocabulary size of Llama3 is much larger than that of the ASR system (about 26 times larger), this causes higher memory consumption for the neural transducer model and slows down the training speed. 
Moreover, using the smaller vocabulary size and the tokenizer trained from ASR data provide performance benefits, with 
8.6\%
WERR on test-clean.
Therefore, the vocabulary adaptation allows the LLM to be more flexibly integrated into speech systems.

\subsubsection{Multi-lingual LibriSpeech (MLS) results}
Experiments were also conducted on the large-scale MLS data, and the conclusions are generally consistent with those on LibriSpeech: Transducer-Llama can still fully utilize the LLMs on online multilingual ASR, and the weak-to-strong LM swap approach is effective. For example, Transducer-Llama outperformed the strong FT baseline with 8.7\% WERR when the LSTM LM was used as the non-blank predictor. When the Llama3 was used, 13.8\% WERR was achieved. 
Llama3 performs slightly better than FFT Llama2 on average, 
while its fixed Transformer layers preserve the potential for other tasks.
This section further constructs an offline Transducer-Llama using the offline Conformer encoder (71M) from \cite{10447605}, in order to compare with other model architectures and published results. The proposed offline Transducer-llama demonstrates superior performance over monolingual + LM baselines (8.2\% WERR), and outperforms an offline speech LLM \cite{10447605} by 7.7\% WERR.

\section{Conclusions}
\label{sec:print}
This paper proposes the Transducer-Llama framework, which naturally integrates LLMs into online ASR. With the vocabulary adaptation technique, Transducer-Llama retains the flexibility of using the tokenizer specifically designed for ASR systems. By using the proposed weak-to-strong LM swap strategy, the LLM can be fully utilized to boost ASR performance.
Moreover, with the vocabulary adaptation and a weak LM used during training, Transducer-Llama can maintain good training speed. In addition, MWER training further improves Transducer-Llama when using an LLM.
Experiments on LibriSpeech and Multi-lingual LibriSpeech (MLS) data show that the proposed Transducer-Llama gave a 17\% relative WER reduction (WERR) over a strong FT baseline and 32\% WERR over an RNN-T baseline.


\bibliographystyle{ieeetr}
\bibliography{strings,refs}

\end{document}